\pdfoutput=1

\documentclass[11pt]{article}

\usepackage{acl}

\usepackage{times}
\usepackage{latexsym}
\usepackage{amssymb}
\usepackage{multicol}
\usepackage{multirow}
\usepackage{CJKutf8}
\usepackage{xcolor}
\usepackage{booktabs}

\usepackage[T1]{fontenc}

\usepackage[utf8]{inputenc}

\usepackage{microtype}

\usepackage{inconsolata}

\usepackage{graphicx}
\usepackage{amsmath}

\title{Mechanistic Understanding and Mitigation of Language Confusion \\in English-Centric Large Language Models}

\author{Ercong Nie$^{1,2}$ \qquad Helmut Schmid$^{1}$ \qquad Hinrich Sch\"utze$^{1,2}$ \\
$^{1}$Center for Information and Language Processing (CIS), LMU Munich, Germany \\
$^{2}$ Munich Center for Machine Learning (MCML), Germany \\
\texttt{nie@cis.lmu.de}}

\begin{document}
\maketitle
\begin{abstract}
Language confusion---where large language models (LLMs) generate unintended languages against the user's need---remains a critical challenge, especially for English-centric models. We present the first mechanistic interpretability (MI) study of language confusion, combining behavioral benchmarking with neuron-level analysis. Using the Language Confusion Benchmark (LCB), we show that confusion points (CPs)---specific positions where language switches occur---are central to this phenomenon. Through layer-wise analysis with TunedLens and targeted neuron attribution, we reveal that transition failures in the final layers drive confusion. We further demonstrate that editing a small set of critical neurons, identified via comparative analysis with a multilingual-tuned counterpart, 
substantially mitigates confusion while largely preserving general competence and fluency. 
Our approach matches multilingual alignment in confusion reduction for many languages and yields cleaner, higher-quality outputs. These findings provide new insights into the internal dynamics of LLMs and highlight neuron-level interventions as a promising direction for robust, interpretable multilingual language modeling. Code and data are available at: \url{https://github.com/ercong21/lang_confusion}.

\end{abstract}

\section{Introduction}

\begin{figure}[t]
    \centering
    \includegraphics[width=.875\linewidth]{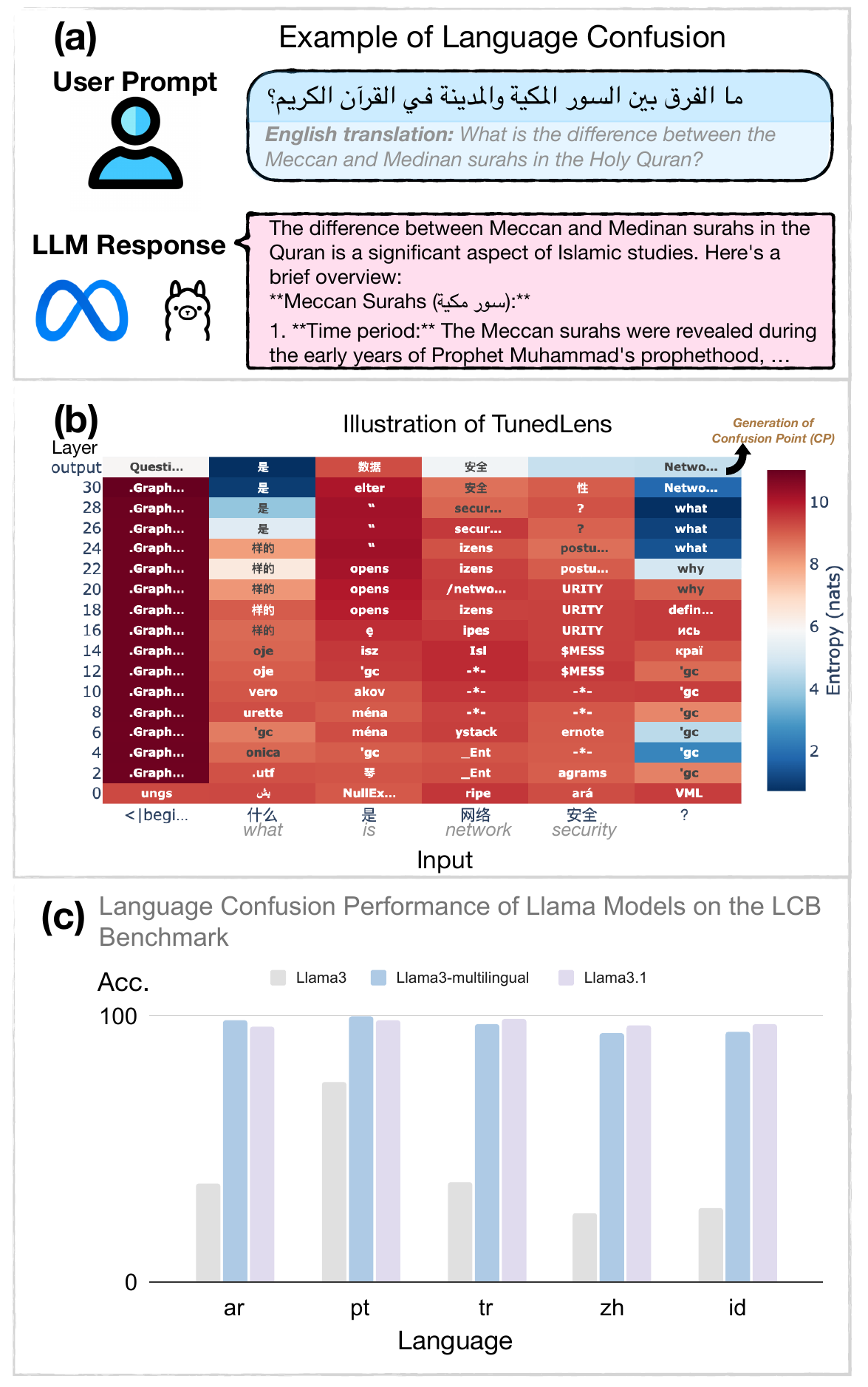}
    \caption{Language Confusion in LLMs. (a) An example of the language confusion phenomenon. (b) Visualization of internal model dynamics using TunedLens, highlighting how the confusion point emerges during generation. (c) Benchmarking results of three Llama models on the LCB benchmark across 5 languages.
    }
    \label{figure1}
\end{figure}

Current Large Language Models (LLMs), such as GPT-4~\citep{achiam2023gpt}, PaLM 2~\citep{anil2023palm}, and Llama 3~\citep{grattafiori2024llama}, have demonstrated exceptional linguistic competence across a wide range of complex tasks that require abstract knowledge and reasoning~\citep{dong-etal-2024-survey,wei2022chain,nie-etal-2025-bmike,wang2025language}. 
Early LLMs were predominantly trained on massive amounts of English text data, with some limited exposure to other languages, resulting in initially constrained multilingual capabilities~\citep{touvron2023llama}. Recent advances, such as multilingual continued pretraining and instruction tuning, have substantially extended these models' ability to support multiple languages~\citep{zhu2023extrapolating,tan-monz-2023-towards,shaham-etal-2024-multilingual,kew-etal-2024-turning,wang-etal-2025-language}.
As a result, contemporary English-centric LLMs have become foundational tools for multilingual communication, multilingual content generation, and cross-lingual applications~\citep{nie-etal-2023-cross,bang-etal-2023-multitask,ahuja-etal-2023-mega,li-etal-2023-crosslingual,asai-etal-2024-buffet}. 
However, despite their impressive capabilities, a persistent and underexplored limitation remains: LLMs can fail to generate text in the user’s intended language, even when explicitly instructed---a phenomenon termed language confusion~\citep{marchisio-etal-2024-understanding}. 
Language confusion manifests as full-response, line-level, or word-level switches into unintended languages, severely undermining user experience and model reliability, especially for non-English speakers (Figure \ref{figure1}a).

Recent work by ~\citet{marchisio-etal-2024-understanding} provides the first systematic characterization of language confusion, introducing the Language Confusion Benchmark (LCB) and associated metrics to quantify this phenomenon across a diverse set of languages and models. Their evaluation revealed that even state-of-the-art LLMs are susceptible to language confusion, with English-centric LLMs such as Llama2, Llama3, and Mistral exhibiting particularly high rates of unintended language switching, especially in the absence of targeted multilingual alignment (Figure \ref{figure1}c). While ~\citet{marchisio-etal-2024-understanding} propose several mitigation strategies, including decoding adjustments, prompting techniques, and multilingual fine-tuning, these approaches remain largely surface-level, offering limited insight into the internal mechanisms that give rise to language confusion.

A key observation from prior work is the identification of confusion points---specific positions in the generation process where the model abruptly switches to an unintended language. However, the model's internal dynamics leading to these confusion points and their causal role in language confusion remain largely unexplored. This gap is particularly salient given the parallels to human bilingual code-switching, where switch points between languages are cognitively significant as extensively studied in psycholinguistics~\citep{solorio-liu-2008-learning,bullock2009cambridge}. Further discussions on the relationship between code-switching in human communication and language confusion in LLM text generation are provided in Appendix \ref{further_discussion}.

In this work, we move beyond behavioral evaluation to open the black box of LLMs, leveraging mechanistic interpretability (MI) methods~\citep{arthur2023automated,rai2024practical,saphra-wiegreffe-2024-mechanistic,sharkey2025open} to investigate the internal representations and neuron-level processes underlying language confusion. We first empirically demonstrate that confusion points are critical drivers of language confusion: targeted replacement at these points can substantially reduce confusion across languages. Building on the CP token, we employ MI tools such as \emph{TunedLens}~\citep{belrose2023eliciting} to trace the evolution of language representations through the model’s layers, revealing that confusion typically arises from transition failures in the final layers, where latent conceptual representations are mapped to surface forms in the target language (Figure \ref{figure1}b).
To further elucidate the mechanism, we conduct a neuron-level analysis, identifying specific neurons in the last layers whose activity is predictive of successful or failed language transitions at confusion points. Inspired by recent advances in neuron attribution and editing, we show that targeted manipulation of only 100 neurons can effectively mitigate language confusion, offering a novel, model-internal approach to improving multilingual reliability.
Our findings provide the first mechanistic account of language confusion in LLMs, bridging the gap between behavioral benchmarks and internal model dynamics. By highlighting the central role of confusion points and their neural mechanisms, we lay the groundwork for more robust, interpretable, and cognitively informed multilingual language models.

Our work makes the following contributions: (1) We provide the first mechanistic interpretability study of language confusion in English-centric LLMs, revealing the central role of confusion points in unintended language switching; (2) We employ layer-wise and neuron-level analyses to trace the internal dynamics leading to language confusion and identify critical late-layer neurons responsible for transition failures; (3) We propose and validate a principled neuron selection and editing strategy that effectively mitigates language confusion and preserves the model’s general competence and output quality.

\begin{table*}[t]
    \centering
    \scalebox{0.83}{
    \begin{tabular}{clll}
    \toprule
     \textbf{Dataset} & \textbf{Data Source} & \textbf{Language} & \textbf{Prompt Example} \\
     \hline
      Aya   &  \multirow{2}{*}{Human-generated} & \multirow{2}{*}{ar, en, pt, tr, zh} & 
\begin{CJK*}{UTF8}{gbsn}
请简单介绍诗人李白的背景。
\end{CJK*}  \\
\citep{singh-etal-2024-aya} & & & \textcolor{gray}{\textit{Briefly introduce the poet Li Bai.}} \\
\hline
      Dolly & \multirow{2}{*}{MT post-edited} & \multirow{2}{*}{ar, es, fr, hi, ru} & Qu'est-ce qui est plus important, l’inné ou l’acquis?\\
     \citep{singh-etal-2024-aya} & & & \textcolor{gray}{\textit{What is more important, nature or nurture?}}\\
     \hline
      Native & \multirow{2}{*}{Human-generated} & \multirow{2}{*}{es, fr, ja, ko} & \begin{CJK}{UTF8}{mj} 콘크리트는 뭘로 만든거야?
      \end{CJK}\\
     \citep{marchisio-etal-2024-understanding} & & & \textcolor{gray}{\textit{What is concrete made of?}}\\
     \hline
      Okapi & \multirow{2}{*}{Synthetic + MT} & ar, en, pt, zh,it, & Schreib einen Aufsatz von 500 Wörtern zum Thema KI.\\
     \citep{lai-etal-2023-okapi} & & fr, de, id, es, vi & \textcolor{gray}{\textit{Write a 500-word essay on AI.}} \\
\bottomrule
    \end{tabular}}
    \caption{Overview and Prompt Example of the LCB Benchmark (monolingual part). The number of examples per language is 100 in each dataset.}
    \label{lcb}
\end{table*}

\section{Related Work} 

\paragraph{Mechanistic Interpretability Methods}
Mechanistic interpretability (MI) seeks to reverse-engineer neural networks by decomposing their computations into human-understandable components~\citep{stolfo-etal-2023-mechanistic,wang-etal-2024-unveiling,men-etal-2024-unlocking}. A central technique in MI is the projection of intermediate representations into the vocabulary space, as implemented by tools such as LogitLens~\citep{logit-lens} and TunedLens~\citep{belrose2023eliciting}, which enable researchers to track how information and predictions evolve across layers~\citep{dar-etal-2023-analyzing,pal-etal-2023-future}. In addition to layer-wise analysis, recent work has focused on identifying, attributing, and intervening on important neurons---those whose activations are strongly correlated with specific linguistic functions or behaviors~\citep{bau2020understanding,geva-etal-2022-transformer,yu-ananiadou-2024-neuron,tan-etal-2024-neuron}. Methods for neuron selection and editing, as well as circuit-level analysis~\citep{elhage2021mathematical,wang2023interpretability}, have proven effective for uncovering the internal structure underlying phenomena such as factual recall~\citep{meng2022locating,geva-etal-2023-dissecting}, reasoning processing~\citep{yu-ananiadou-2024-interpreting}, and now, as in our work, language confusion. By leveraging these MI techniques, we aim to provide a granular, causal understanding of how and why language confusion arises in multilingual LLMs, and to identify actionable intervention points for mitigation.

\begin{table*}[t]
\centering
\scalebox{0.71}{
\begin{tabular}{ccrrrrrrrrrrrrrrrr} 
\toprule
             \textbf{Model}  &   \textbf{Metric}  & \multicolumn{1}{l}{ar} & \multicolumn{1}{l}{en} & \multicolumn{1}{l}{pt} & \multicolumn{1}{l}{tr} & \multicolumn{1}{l}{zh} & \multicolumn{1}{l}{es} & \multicolumn{1}{l}{fr} & \multicolumn{1}{l}{hi} & \multicolumn{1}{l}{ru} & \multicolumn{1}{l}{ja} & \multicolumn{1}{l}{ko} & \multicolumn{1}{l}{de} & \multicolumn{1}{l}{id} & \multicolumn{1}{l}{it} & \multicolumn{1}{l}{vi} & \multicolumn{1}{l}{\textbf{avg}}  \\ 
\hline
Llama3         & \textit{LPR} & 33.0                    & 99.5                    & 71.0                    & 33.0                    & 19.3                    & 73.0                    & 59.3                    & 8.0                     & 28.0                    & 14.0                    & 23.0                    & 19.0                    & 22.0                    & 34.0                    & 11.0                    & \textbf{36.5}                     \\ 

\textit{(original)}     & \textit{Acc} & 33.7                    & 99.8                    & 74.5                    & 37.5                    & 23.4                    & 77.1                    & 64.1                    & 15.1                    & 28.2                    & 17.1                    & 23.6                    & 23.0                    & 27.3                    & 39.8                    & 14.8                    & \textbf{39.9}                     \\ 
\hline
Llama3         & \textit{LPR} & 71.0                    & 99.0                    & 93.0                    & 50.0                    & 57.3                    & 94.3                    & 84.0                    & 37.0                    & 78.6                    & 50.0                    & 45.0                    & 60.0                    & 67.0                    & 86.0                    & 62.0                    & \textbf{68.9}                   \\ 

\textit{(replace)}      & Acc & 74.8                    & 99.6                    & 95.4                    & 55.5                    & 64.1                    & 95.3                    & 86.5                    & 47.6                    & 83.1                    & 55.3                    & 48.6                    & 62.3                    & 77.7                    & 87.5                    & 66.1                    & \textbf{73.3}                      \\ 
\hline
Llama3         & \textit{LPR} & 98.3                    & 98.5                    & 99.0                    & 95.8                    & 88.8                    & 98.3                    & 95.9                    & 97.0                    & 100.0                   & 93.5                    & 100.0                   & 100.0                   & 88.8                    & 100.0                   & 97.9                    & \textbf{96.8}                      \\ 

\textit{(multilingual)} & \textit{Acc} & 98.7                    & 99.5                    & 99.8                    & 96.9                    & 93.8                    & 99.3                    & 96.9                    & 97.5                    & 100.0                   & 95.8                    & 100.0                   & 100.0                   & 94.2                    & 100.0                   & 97.9                    & \textbf{98.0}                     \\
\bottomrule
\end{tabular}}
\caption{Impact of Confusion Point Replacement on Language Confusion Metrics. Line-level pass rate (LPR) and line-level accuracy for original Llama3-8B, multilingual Llama3-8B, and Llama3-8B with confusion point replacement, reported by language.}
\label{cp_replace}
\end{table*}

\paragraph{Multilingual Interpretability}
Recent research has begun to probe the internal representations of English-centric and multilingual LLMs to understand how they process and transfer information across languages~\citep{he2024large,zhao2024llama,kojima-etal-2024-multilingual,zhang-etal-2024-respond}.~\citet{wendler-etal-2024-llamas} show that models like Llama2 often rely on English as an internal pivot language and can disentangle language and conceptual representations in controlled tasks. \citet{fierro2025multilingual} examine how mechanisms identified in monolingual contexts generalize to multilingual settings. \citet{wang-etal-2025-lost-multilinguality} investigate the internal causes of crosslingual factual inconsistencies, revealing how multilingual models transition from language-independent to language-specific processing. However, prior work has not systematically connected these internal mechanisms to language generation errors such as language confusion.

\section{Revisiting Language Confusion: Benchmark Insights}

\subsection{Recap of Language Confusion Benchmark}
The Language Confusion Benchmark (LCB)~\citep{marchisio-etal-2024-understanding} provides a systematic framework for evaluating the ability of LLMs to generate text in the user’s intended language. 
The benchmark covers 15 typologically diverse languages and uses a diverse set of prompts sourced from human-written, post-edited, and synthetic datasets to evaluate models, ensuring coverage of a wide range of domains and linguistic structures (Table \ref{lcb}). 
In this work, we focus on the \textit{monolingual setting} of LCB, where the prompt and expected response are in the same language.
This setting is particularly relevant for mechanistic interpretability research, as it isolates language confusion phenomena from the additional complexities of explicit cross-lingual transfer.

To quantify language confusion, we adopt two key metrics from LCB: \textbf{line-level pass rate (LPR)} and \textbf{line-level language accuracy (Acc)}. LPR measures the percentage of model responses in which every line is in the correct language. Acc reflects the proportion of individual lines across all responses that are correctly generated in the target language. Both metrics rely on automatic language identification using the fastText classifier~\citep{joulin2016fasttext,joulin-etal-2017-bag}, which efficiently detects the language of each line in the generated output.

We conducted preliminary benchmarking experiments on LCB with three instruction-tuned LLMs: \textit{\textbf{Llama3-8B}} \textit{(English-centric, no multilingual instruction tuning)}, \textbf{\textit{Llama3-8B-multilingual}} \textit{(multilingually instruction-tuned)}~\citep{devine-2024-tagengo}, and \textbf{\textit{Llama3.1-8B}} \textit{(multilingually optimized)}.
As shown in Figure \ref{figure1}c,
\textit{Llama3-8B} exhibits substantial language confusion, with frequent line-level switches to unintended languages (mostly English). In contrast, both \textit{Llama3-8B-multilingual} and \textit{Llama3.1-8B} achieve near-perfect line-level accuracy, demonstrating the effectiveness of multilingual instruction tuning and targeted optimization for multilingual dialogue.

Given these findings, our work centers on understanding and mitigating the language confusion observed in English-centric \textit{Llama3-8B}. By leveraging mechanistic interpretability methods, we aim to uncover the internal causes of confusion and develop interventions that can bring its performance closer to that of explicitly multilingually-tuned models. In the following subsection, we delve deeper into the significance of confusion points as critical junctures in the generation process.

\subsection{Significance of Confusion Points}
A confusion point (CP) is the position in a model’s output where the first token of an unintended language abruptly appears, marking the onset of language confusion~\citep{marchisio-etal-2024-understanding}. This concept is inspired by psycho- and neurolinguistic research on code-switching, where the precise location of a language switch---known as a switch point---is central to understanding bilingual language production and processing~\citep{blanco2017bilingual,suurmeijer2020structural}. 
To empirically assess the role of CPs in LLM language confusion, we conduct a \emph{replacement experiment} on \textit{Llama3-8B}. For each instance of language confusion, we identify the CP using the fastText language detector. We then replace the token at the subword-level CP with the corresponding token generated by \textit{Llama3-8B-multilingual}, which achieves near-perfect language accuracy, under the same prompt. This approach is motivated by the psycholinguistic observation that, in human code-switching, the choice at the switch point strongly influences the subsequent language trajectory~\citep{moreno2002switching,lai2020examining}. 

\begin{figure*}[t]
    \centering
    \includegraphics[width=\linewidth]{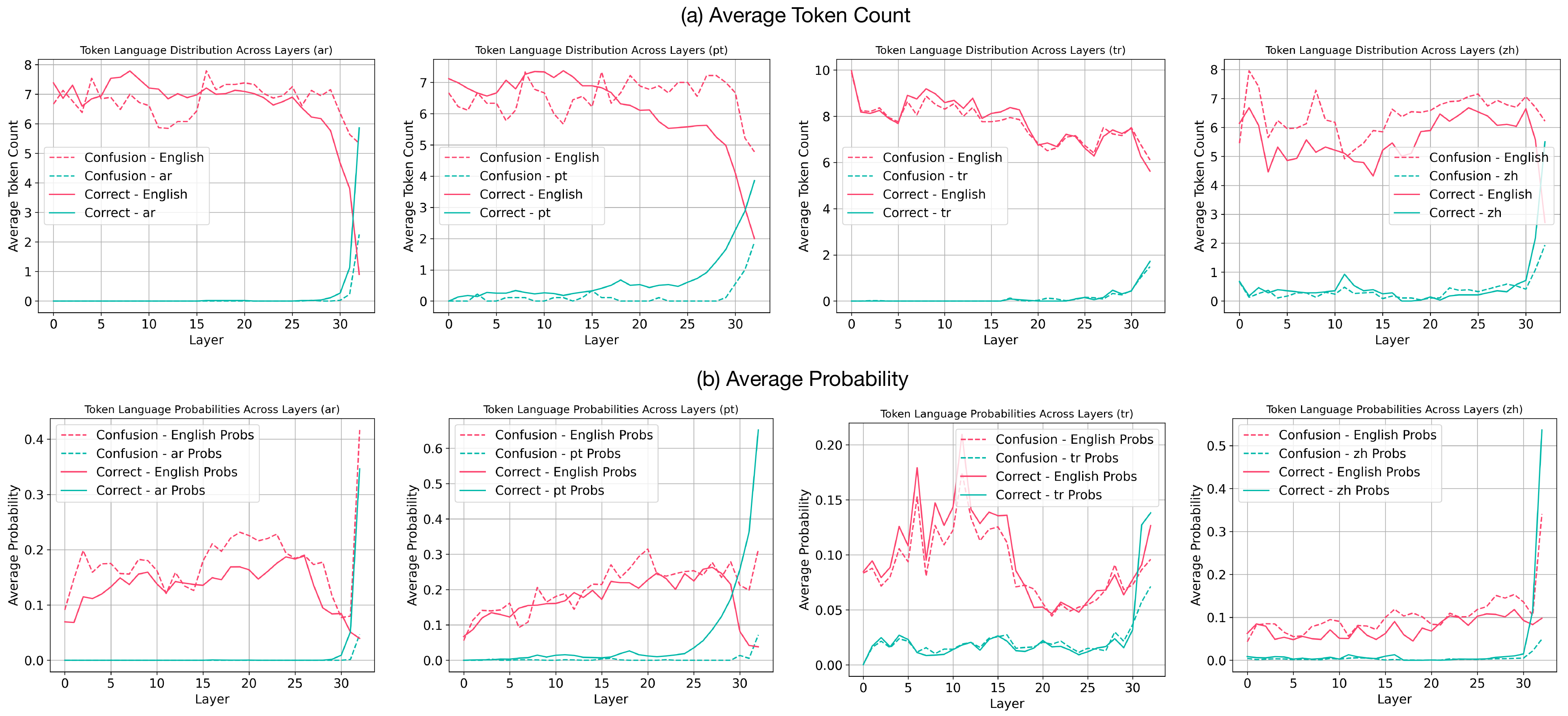}
    \caption{Average token counts and probabilities for English and target language tokens among the top-10 predictions at each layer, shown for both correct and confusion samples across four languages from \textit{Aya}.}
    \label{tunedlens}
\end{figure*}

Our results, summarized in Table \ref{cp_replace},
show a substantial reduction in language confusion after CP replacement, even though our method does not represent an oracle upper bound. 
These findings highlight the centrality of confusion points in the emergence of language confusion and motivate us to base the subsequent mechanistic analysis and targeted interventions on the generation of the token at the CP position.

\section{Mechanistic Analysis of Language Confusion Points}

\subsection{Analyzing Layer-wise Language Transition}
A central question in understanding language confusion is where and how the model’s internal representations fail to transition from a shared conceptual space to the intended target language. Motivated by recent findings that English-centric LLMs process information in a latent, often English-biased, conceptual space before converting it to the target language in the final layers~\citep{wendler-etal-2024-llamas,wang-etal-2025-lost-multilinguality}, we conduct a detailed layer-wise analysis of this transition using TunedLens~\citep{belrose2023eliciting}.

We employ TunedLens, the more reliable variant of LogitLens~\citep{logit-lens}, to unembed the hidden states of \textit{Llama3-8B} at each layer into the vocabulary space. With this, we inspect every layer of the model and extract the top 10 predicted tokens with the largest logits at the position immediately preceding the confusion point (CP) (for confusion cases) or the output token (for correct cases). For each layer, we compute the average number and summed probabilities of English and target language tokens among the top-10 predictions, using fastText for language identification.
Our analysis focuses on four typologically diverse languages (Arabic, Portuguese, Turkish, Chinese) from the LCB benchmark. We separate samples into two groups: (1) \textit{Correct}---where the model generates the intended language throughout, and (2) \textit{Confusion}---where the model switches to an unintended language at a CP. For confusion samples, we analyze the model’s state up to the token before the CP.

Figure \ref{tunedlens} presents the evolution of language token counts and probabilities across layers for both groups. In early and middle layers, English tokens dominate the top-10 predictions for all languages, reflecting the English-centric latent conceptual space of \textit{Llama3-8B}. This is consistent with prior work showing that LLMs encode information in a shared, language-agnostic space in intermediate layers.
In the final layers, a sharp transition emerges. For correct samples, the number and probability of target language tokens rise steeply, overtaking English tokens in the last few layers---indicating a successful transition to the target language surface form. In contrast, for confusion samples, this transition fails: English tokens remain dominant or even increase, while target language tokens lag behind. This failure to shift from the latent conceptual space to the target language at the critical moment leads to CPs and erroneous output.

Our layer-wise analysis with TunedLens reveals that the transition to the target language occurs in the final layers, and that failures in this process are tightly linked to language confusion. These findings provide direct evidence that language confusion in \textit{Llama3-8B} is primarily caused by transition failures in the last few layers, motivating our subsequent neuron-level investigation to pinpoint and intervene on the specific components responsible for these failures.

\subsection{Localizing Critical Neurons at Confusion Points}
A key step toward understanding and mitigating language confusion is to identify which neurons are most responsible for the emergence of confusion points. Building on recent advances in neuron-level attribution~\citep{geva-etal-2022-transformer,yu-ananiadou-2024-neuron}, we adopt a static, efficient method to locate and analyze the most influential feed-forward network (FFN) neurons in \textit{Llama3-8B}.
\paragraph{Methodology}
In the inference pass in decoder-only LLMs, for a given input sequence, each layer output $h_i^l$ (layer $l$, token position $i$) is a sum of the previous layer's output $h_i^{l-1}$, the attention output $A_i^l$, and the FFN output $F_i^l$:
\begin{equation}
    h_i^l=h_i^{l-1}+A_i^l+F_i^l
\end{equation}
The FFN output $F_i^l$ is calculated by a non-linear $\sigma$ on two MLPs $W^l_{fc1}\in \mathbb{R}^{N\times d}$ and $W^l_{fc2}\in \mathbb{R}^{d\times N}$:
\begin{equation}
    F_i^l = W^l_{fc2}\sigma(W^l_{fc1}(h_i^{l-1}+A_i^l))
\end{equation}
Following ~\citet{geva-etal-2021-transformer}, the FFN layer output $F_i^l$ can be represented as a weighted sum over neuron subvalues:
\begin{equation}
    F_i^l=\sum_{k=1}^N m_{i,k}^l \cdot fc2^l_k
\end{equation}
\begin{equation}
    m_{i,k}^l=\sigma(fc1^l_k \cdot (h_i^{l-1}+A_i^l))
\end{equation}
where $fc2^l_k$ is the $k$-th column of $W^l_{fc2}$, and $m_{i,k}^l$ is derived from the inner product between the residual output $(h_i^{l-1}+A_i^l)$ and $fc1^l_k$, the $k$-th row of $W^l_{fc1}$.

\citet{geva-etal-2022-transformer} and \citet{dar-etal-2023-analyzing} project FFN neuron subvalues with unembedding matrices to compute the token probability distribution. 
To quantify the importance of each neuron for generating a specific token (e.g., at a confusion point), we adopt the log probability increase method of~\citet{yu-ananiadou-2024-neuron}. 
For a neuron in the $l$-th FFN layer $v^l$, its importance score is defined as the increase in log probability of the target token when $v^l$ is added to the residual stream $A^l+h^{l-1}$, compared to the baseline without $v^l$:
\begin{equation}
\begin{split}
    Imp(v^l)=\log(p(w|v^l+A^l+h^{l-1})- \\\log(p(w|A^l+h^{l-1})
\end{split}
\end{equation}
This approach efficiently identifies neurons whose activations most strongly influence the model’s prediction at a given position.

\paragraph{Experimental Observations}
\begin{figure}[h]
    \centering
    \includegraphics[width=.9\linewidth]{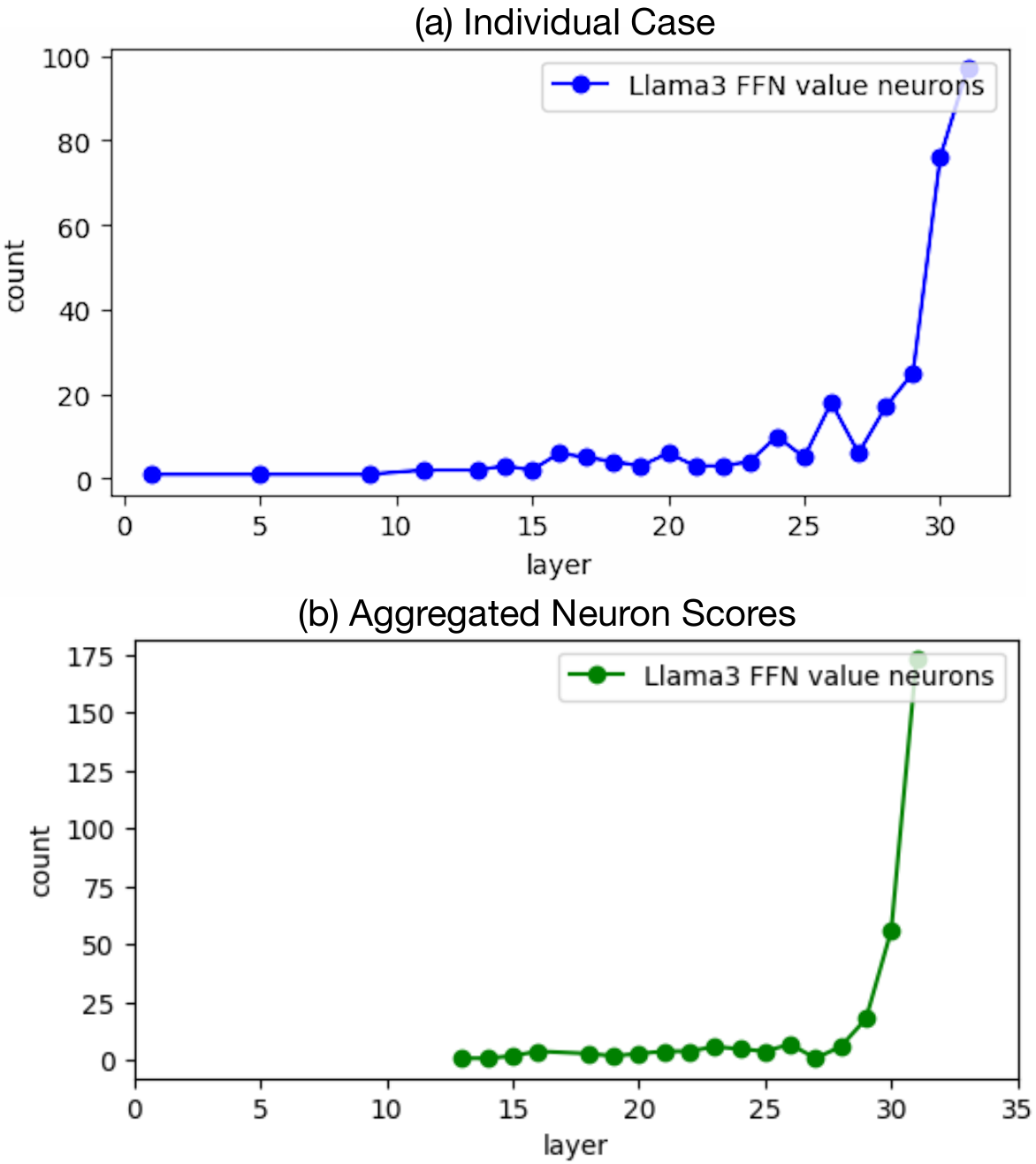}
    \caption{Distribution of Important Neurons Associated with Confusion Points in \textit{Llama3-8B}. (a) Distribution of the top 300 most important FFN neurons across layers for an individual Chinese prompt ``\begin{CJK*}{UTF8}{gbsn}请解释拆东墙补西墙的意思。\end{CJK*}\textcolor{gray}{\textit{(Please explain `\begin{CJK*}{UTF8}{gbsn}拆东墙补西墙\end{CJK*}.')}}'' from Aya. (b) Aggregated distribution of important neuron scores across all Chinese test samples in Aya.}
    \label{distribution}
\end{figure}
We apply this method to \textit{Llama3-8B} on confusion samples from the LCB benchmark, focusing on the token position immediately preceding each confusion point. For each sample and language, we compute the importance scores for all 14,336 FFN neurons in each layer of \textit{Llama3-8B}, rank them, and select the top 300 most important neurons per sample. 
We then analyze the distribution of these critical neurons across layers, both for individual samples and aggregated over all samples in a language.
Our analysis reveals a striking concentration of important neurons in the final layers, as visualized in Figure \ref{distribution}.
This pattern holds both at the single-sample level and when aggregating across samples, indicating that the emergence of confusion points is primarily driven by late-layer FFN activity. 
We further rank neurons by their frequency of appearance in the top 300 sets across samples, finding that a subset of neurons consistently recurs as highly influential for confusion points.

\begin{table*}[t]
\centering
\scalebox{0.7}{
\begin{tabular}{llllllllllllllll} 
\toprule
            & ar    & pt    & tr    & zh    & es    & fr    & hi    & ru    & ja    & ko    & de    & id    & it    & vi    & \textbf{Avg.}    \\ 
\hline
\textit{original}    & 33.44 & 74.26 & 37.55 & 24.04 & 77.15 & 63.16 & 16.47 & 28.20 & 17.44 & 23.50 & 23.00 & 27.33 & 39.83 & 14.79 & \textbf{35.73}  \\ 

\textit{freq }       & 31.75 & 75.10 & 36.51 & 22.09 & 76.29 & 66.98 & 18.66 & 27.70 & 19.29 & 23.08 & 22.25 & 27.83 & 39.45 & 13.58 & \textbf{35.75}  \\ 

\textit{score}       & 76.97 & 93.41 & 67.61 & 80.63 & 91.22 & 74.77 & 60.00 & 50.32 & 53.50 & 33.25 & 40.27 & 53.58 & 96.00 & 67.56 & \textbf{67.08}  \\ 

\textit{comparative} & 85.45 & 97.12 & 57.27 & 89.39 & 92.20 & 83.17 & 82.74 & 89.43 & 49.95 & 40.33 & 80.82 & 78.94 & 95.25 & 66.50 & \textbf{77.75 }\\
\bottomrule
\end{tabular}}
\caption{Confusion mitigation performance of different selection strategies. Line-level accuracy is reported.}
\label{selection}
\end{table*}

\begin{figure}[h]
    \centering
    \includegraphics[width=\linewidth]{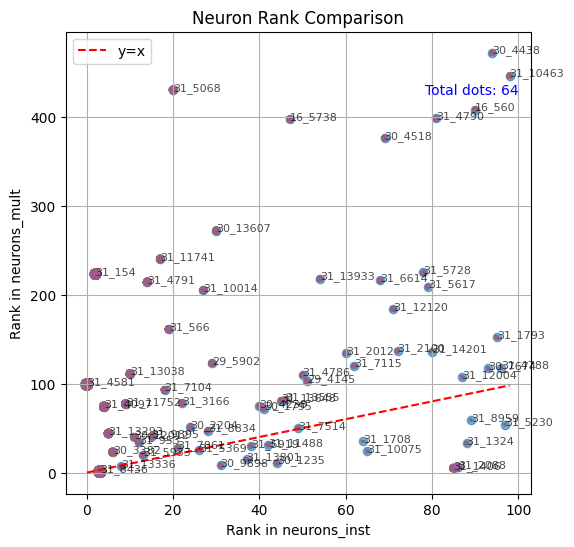}
    \caption{Neuron rank comparison between original Llama3 and multilingual Llama3. Results of Chinese test samples in Aya.}
    \label{comparison}
\end{figure}

To understand the effect of multilingual alignment, we repeat the analysis on \textit{Llama3-8B-multilingual} using the same set of prompts. After multilingual instruction tuning, language confusion is nearly eliminated. Comparing neuron importance scores between the two models (Figure \ref{comparison}), we observe that most neurons critical for confusion in the \textit{Llama3-8B} become much less important in its multilingual counterpart, suggesting that multilingual alignment suppresses the activity of confusion-inducing neurons.
However, a small number of neurons remain important or even increase in importance, likely reflecting their role in encoding general semantic information rather than language-specific transitions.

These findings reinforce the conclusion from our layer-wise analysis: language confusion is tightly linked to the activity of specific FFN neurons in the \textit{final} layers. The suppression of these neurons through multilingual alignment provides a mechanistic explanation for the effectiveness of such tuning. Moreover, the identification of a small set of persistent, semantically important neurons suggests that \textit{targeted} neuron-level interventions could mitigate confusion without harming overall model performance. These insights directly inform our subsequent strategies for neuron-based mitigation of language confusion.

\section{Mitigating Language Confusion via Neuron Editing}
A central challenge in mitigating language confusion via neuron editing is to identify a set of neurons whose intervention effectively reduces confusion without degrading the model’s general competence or fluency. Insights from our previous mechanistic analysis indicate that language confusion is primarily driven by a subset of late-layer FFN neurons. However, indiscriminate deactivation of important neurons risks harming the model’s overall performance. Thus, a principled neuron selection strategy is essential.

\begin{table}[h]
\centering
\scalebox{0.65}{
\begin{tabular}{lcccccc} 
\toprule
         & \multicolumn{1}{c}{token\_num} & \multicolumn{1}{c}{token\_prob} & \multicolumn{1}{c}{fluency} & \multicolumn{1}{c}{acc\_ood} & \multicolumn{1}{c}{xnli} & \multicolumn{1}{c}{senti}  \\ 
\hline
Original & 1.96                            & 24.5                             & 25.8                         & 39.9                          & 46.4                                & 98.4                                  \\ 

Edited   & 3.43                            & 36.8                             & 21.8                         & 74.25                         & 44.9                                & 98.2                                  \\ 

Diff     & 1.47                            & 12.3                             & -4.0                         & 34.4                          & -1.5                                & -0.2                                  \\
\bottomrule
\end{tabular}}
\caption{Results of generalization and robustness of neuron editing. Average performance across languages is reported. Detailed results in Appendix \ref{full_results}.}
\label{robustness}
\end{table}

\subsection{Neuron Selection and Intervention}
We compare three neuron selection strategies: 
(1) \textit{Frequency-Based Selection:} Selects the neurons most frequently identified as important across all confusion samples for a given language. 
(2) \textit{Aggregate Importance Selection:} Ranks neurons by the sum of their importance scores across all confusion samples, selecting those with the highest cumulative influence. 
While this method captures the overall impact, it may still include neurons essential for general language competence.
(3) \textit{Comparative Importance Selection:} Inspired by \citet{yu-ananiadou-2024-interpreting}, this strategy identifies neurons whose importance scores for confusion points decrease most substantially after multilingual alignment. Specifically, for each neuron, we compute the difference in importance score between original \textit{Llama3-8B} and \textit{Llama3-8B-multilingual} on the same input. Neurons with the largest drop are prioritized for intervention, as they are likely to be specifically implicated in language confusion rather than general semantic processing.

For each strategy, we select the top 100 neurons and intervene by setting their activations to zero during generation. We evaluate the impact of each method on the LCB benchmark. Our results (Table \ref{selection}) demonstrate that Comparative Importance Selection achieves the most effective reduction in language confusion, substantially outperforming both frequency-based and aggregate importance methods. Frequency-based selection yields minimal benefit, while aggregate importance provides moderate improvement but still lags behind our proposed \emph{comparative} approach. Notably, the comparative strategy selectively targets neurons implicated in confusion, minimizing collateral impact on general model competence.

\subsection{Generalization and Robustness of Neuron Editing}

To further validate the effectiveness and safety of our Comparative Importance Selection strategy, we conduct a comprehensive evaluation across multiple metrics and experimental setups. Our goal is to ensure that neuron editing not only mitigates language confusion but also preserves the model’s general competence, fluency, and robustness across domains (Table \ref{robustness}).
Details of the experimental setup for the generalization and robustness experiments are provided in Appendix \ref{experiment_details}.

\paragraph{Language Confusion Mitigation}
We first assess the impact of neuron editing on language confusion using the LCB benchmark. In addition to standard metrics (line-level pass rate and line-level accuracy), we analyze the internal output distributions by reporting (1) the number of target language tokens among the top-10 candidates in the final output token logit, and (2) the total probability mass assigned to target language tokens in the top-10. These metrics provide a deeper view of how neuron editing shifts the model’s internal preference toward the intended language, beyond surface-level accuracy.

\paragraph{Robustness on General Tasks}
To evaluate whether neuron editing affects the model’s general capabilities, we test the edited model on widely used multilingual benchmarks, including XNLI~\citep{conneau-etal-2018-xnli} and multilingual sentiment analysis~\citep{keung-etal-2020-multilingual}. We also assess output fluency by measuring the perplexity of generated responses using the multilingual model \texttt{facebook/xglm-564M}~\citep{lin-etal-2022-shot}. Across all these metrics, the edited model maintains performance comparable to the original \emph{Llama3-8B}, indicating that our neuron intervention method does not degrade general language understanding or generation quality.

\paragraph{Out-of-Domain Generalization}
We further examine the generalization of neuron editing by applying neurons selected from one data source (e.g., Aya) to out-of-domain test sets (e.g., Okapi) for the same language. The edited model continues to demonstrate strong mitigation of language confusion, suggesting that the identified neurons capture robust, domain-independent mechanisms underlying confusion points.

\subsection{Comparison with Multilingual Alignment}
To contextualize the effectiveness of neuron editing, we compare the performance of the edited \textit{Llama3-8B} model with that of the multilingual-tuned \textit{Llama3-8B}. Quantitative results show that neuron editing yields confusion mitigation comparable to multilingual instruction tuning for many languages, despite requiring no additional data or retraining. This demonstrates that targeted neuron intervention can match the benefits of extensive multilingual instruction tuning for confusion reduction. However, for languages with limited pretraining exposure, multilingual supervised fine-tuning can achieve stronger performance.

\begin{figure}[h]
    \centering
    \includegraphics[width=.9\linewidth]{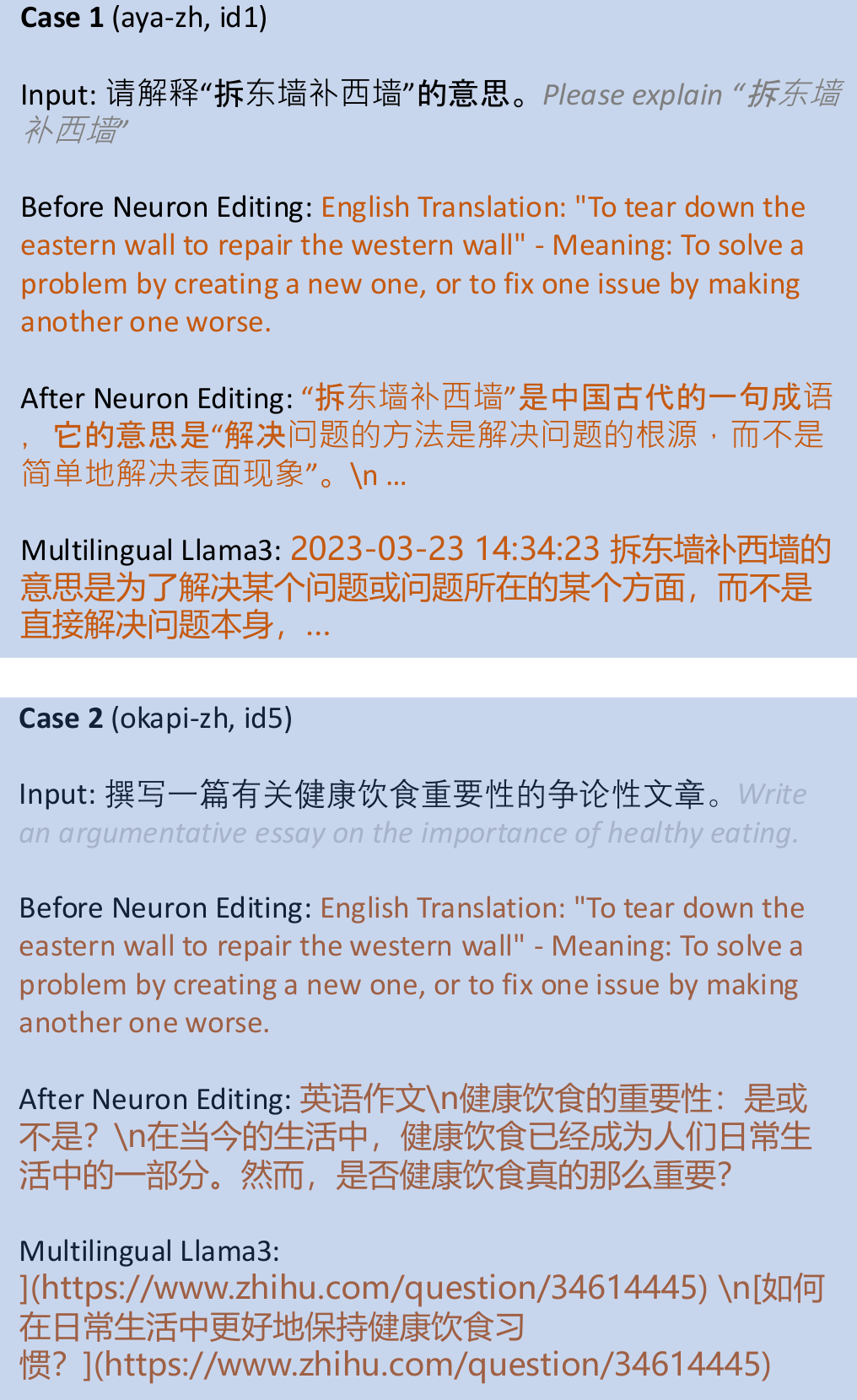}
    \caption{Case study of neuron editing.}
    \label{case}
\end{figure}

Beyond aggregate metrics, qualitative analysis reveals further advantages of the neuron editing approach. Through case studies, we observe that the edited \emph{Llama3-8B} not only generates fluent and accurate responses in the intended target language, but also avoids certain artifacts introduced by multilingual alignment. For example, in several instances (Figure \ref{case}), the multilingual \emph{Llama3-8B} produces outputs containing HTML tags or formatting patterns reminiscent of its instruction tuning data, reflecting the influence of imperfect or noisy multilingual datasets. In contrast, the neuron-edited model consistently produces clean, well-structured, and contextually appropriate responses, free from such extraneous formatting.

These findings highlight a key strength of mechanistic neuron editing: it directly addresses the internal causes of language confusion without introducing side effects from large-scale data-driven alignment. By preserving the original model’s semantic competence and output quality, neuron editing offers a more targeted and interpretable solution. This suggests that, beyond traditional multilingual instruction tuning, mechanistic interpretability-driven interventions can provide a promising path toward high-quality, robust multilingual language models.

\section{Conclusions}
This work provides the first mechanistic interpretability account of language confusion in English-centric LLMs. By tracing confusion points to failures in late-layer transitions and localizing the critical neurons responsible, we demonstrate that targeted neuron editing can robustly mitigate language confusion without sacrificing general competence or fluency. Our approach achieves results comparable with multilingual-tuned models for many languages, while preserving cleaner output quality, though multilingual SFT remains stronger for some low-resource cases. These findings highlight the promise of neuron-level interventions for more reliable and interpretable multilingual language modeling.

\section*{Limitations}
While this work provides the first mechanistic interpretability account of language confusion in English-centric LLMs, several limitations remain. Our analysis primarily focuses on the monolingual setting; cross-lingual contexts, which may involve distinct mechanisms and challenges, are left for future research. Additionally, neuron editing interventions are evaluated on selected benchmark tasks and may require further validation across broader domains and model architectures. Lastly, while our approach identifies and mitigates language confusion, fully understanding how these mechanisms interact with other multilingual phenomena warrants further investigation.

\section*{Ethic Statement}
This research was conducted in accordance with
the ACM Code of Ethics. The datasets that we use are publicly available. We have not intended or do not intend to share any Personally Identifiable Data with this paper. Regarding the usage of AI tools, we only use AI models for language refining.

\section*{Acknowledgments}
We want to thank the anonymous reviewers for
their valuable feedback.
This work was supported
by the German Research Foundation (DFG) under grant SCHU 2246/14-1, Munich Center for Machine Learning (MCML), and China Scholarship Council (CSC).

\bibliography{custom}

\newpage

\appendix

\section{Further Discussion on Code-Switching and Language Confusion}
\label{further_discussion}


\paragraph{Code-switching as a Linguistic Phenomenon}
Code-switching, the practice of alternating between languages within a single conversation or utterance, is a well-studied natural phenomenon in bilingualism and psycholinguistics~\citep{gardner2009code}. 
Code-switching is typically intentional, often reflecting speakers’ identities, social relationships, and contextual adaptation~\citep{treffers2009code,yim2021acculturation}. 
In NLP, code-switching has been explored through evaluating model performance on code-switched data for tasks such as sentiment analysis, machine translation, summarization, and language identification~\citep{khanuja-etal-2020-gluecos,dogruoz-etal-2021-survey,winata-etal-2023-decades}. 
Code-switching is a natural, contextually appropriate strategy in human communication, whereas language confusion, on which our work focuses, is an unintended and erroneous switch to an incorrect language in LLMs~\citep{marchisio-etal-2024-understanding}. Though related to code-switching, language confusion is an unnatural phenomenon that arises from model failures rather than communicative intent. 

\paragraph{Language Confusion and Confusion Points in LLMs}
Language confusion has been observed in various multilingual NLP settings, such as ``source language hallucinations'' in zero-shot cross-lingual transfer~\citep{li-murray-2023-zero,pfeiffer-etal-2023-mmt5,chirkova-nikoulina-2024-key} and ``off-target translation'' in machine translation~\citep{sennrich-etal-2024-mitigating}. In LLMs, this manifests as abrupt, unexpected switches to the wrong language during generation, even under explicit instructions. This issue is particularly prevalent in English-centric models lacking robust multilingual alignment~\citep{zhong2024beyond}.
A key concept in recent work is the \emph{confusion point}---the specific position in generation where the model transitions to an unintended language. Inspired by the importance of code-switching points in human bilingualism, confusion points are central to understanding and diagnosing language confusion in LLMs~\citep{tamargo2016examining}. Unlike natural code-switching, these points reflect internal model failures. Recent benchmarks~\citep{marchisio-etal-2024-understanding} systematically characterize confusion points at response, line, and word levels, revealing their widespread impact and motivating deeper mechanistic investigation, as pursued in this work.

\section{Full Experimental Results}
\label{full_results}

Table \ref{benchmarking_result} presents the full benchmarking results.
Table \ref{full_cp_results} shows the full results of the CP replacement experiment.
Tables \ref{token_prob} and \ref{full_generalization} present the full results of robustness and generalization experiments.

\section{Detailed Experimental Setup}
\label{experiment_details}

\subsection{Models} 
We primarily use three variants of the Llama3 family for our experiments: 
\begin{itemize} 
\item \textbf{Llama3-8B}: The baseline English-centric model without multilingual instruction tuning. 
\item \textbf{Llama3-8B-multilingual}: The multilingual instruction-tuned version, as described in \citep{devine-2024-tagengo}. 
\item \textbf{Llama3.1-8B}: An improved model optimized for multilingual dialogue. \end{itemize} All models are used in their publicly released forms unless otherwise stated. For neuron editing experiments, we intervene on \textit{Llama3-8B} using the strategies described in Section 5.

\subsection{Datasets and Tasks} 
\paragraph{Language Confusion Benchmarking and Replacement Experiments} 
We use the Language Confusion Benchmark (LCB) \citep{marchisio-etal-2024-understanding} for all language confusion detection and mitigation experiments. LCB covers 15 typologically diverse languages and comprises several monolingual and cross-lingual datasets: 
\begin{itemize}
\item \textbf{Monolingual sources}: Aya (human-generated), Dolly (post-edited), Native (human-generated), and Okapi (synthetic + machine translated). 
\item \textbf{Languages}: Arabic, English, Portuguese, Turkish, Chinese, Spanish, French, Hindi, Russian, Japanese, Korean, German, Indonesian, Italian, Vietnamese. 
\end{itemize} 
All main benchmarking and confusion point replacement experiments are run on the monolingual portions of LCB, using 100 prompts per language per dataset as described in Table 1.

\paragraph{Robustness and Generalization Experiments} To assess the robustness and generalization of neuron editing, we evaluate on: 
\begin{itemize} 
\item \textbf{XNLI} \citep{conneau-etal-2018-xnli}: Cross-lingual natural language inference in 15 languages. 
\item \textbf{Multilingual Sentiment Analysis}: Standard multilingual sentiment datasets (including German, Spanish, French, Japanese, and Chinese). It is a binary classification task derived from the multilingual Amazon review dataset~\citep{keung-etal-2020-multilingual}. 
\item \textbf{Out-of-domain LCB evaluation}: For each language, neurons are selected from one LCB source (e.g., Aya), then tested on a different source (e.g., Okapi) to assess generalization. 
\end{itemize}

\subsection{Metrics} 
\paragraph{Language Confusion Metrics} We adopt two primary metrics from LCB: 
\begin{itemize} 
\item \textbf{Line-level Pass Rate (LPR)}: Percentage of responses where every line is in the correct language. 
\item \textbf{Line-level Accuracy}: Proportion of lines generated in the correct language. \end{itemize} 
Language identification for these metrics is performed using the fastText classifier \citep{joulin2016fasttext}.

\paragraph{Internal Model Metrics} 
We further report: 
\begin{itemize} 
\item \textbf{Target Language Token Count}: Number of target language tokens among the top-10 output logits in the final layer. 
\item \textbf{Target Language Token Probability}: Total probability mass assigned to target language tokens in the top-10 output logits. 
\end{itemize}

\paragraph{Generalization and Fluency Metrics} \begin{itemize} 
\item 
\textbf{XNLI and Sentiment Accuracy}: Standard classification accuracy on XNLI and multilingual sentiment analysis tasks. 
\item \textbf{Fluency (Perplexity)}: Perplexity of generated outputs, measured using the multilingual \texttt{facebook/xglm-564M} model \citep{lin-etal-2022-shot}. \end{itemize}

\subsection{Implementation Details} All experiments are run on NVIDIA A100 GPUs. Prompt formatting and decoding settings follow the LCB benchmark defaults. Neuron interventions are implemented at inference time via custom hooks in PyTorch, zeroing out selected neuron activations layer-wise as described in Section 5.1. For TunedLens analysis, we use the public implementation from \citet{belrose2023eliciting}.


\begin{table*}
    \centering
    \includegraphics[width=.87\linewidth]{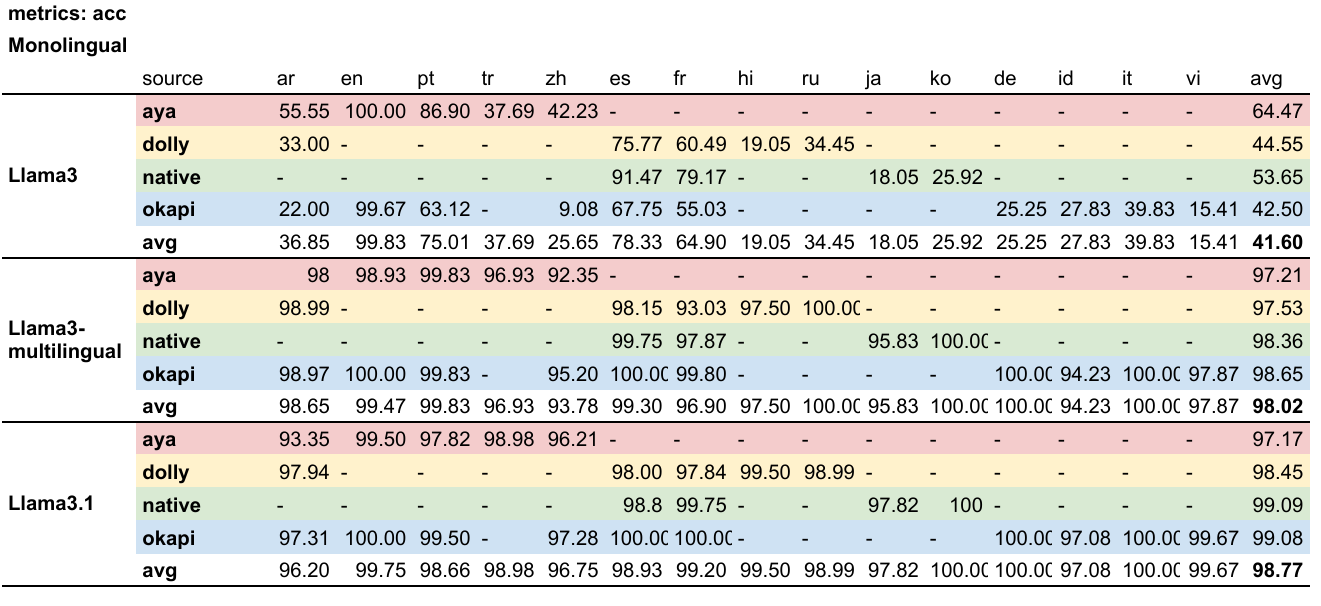}
    \caption{Full benchmarking results on LCB.}
    \label{benchmarking_result}
\end{table*}

\begin{table*}
    \centering
    \includegraphics[scale=1]{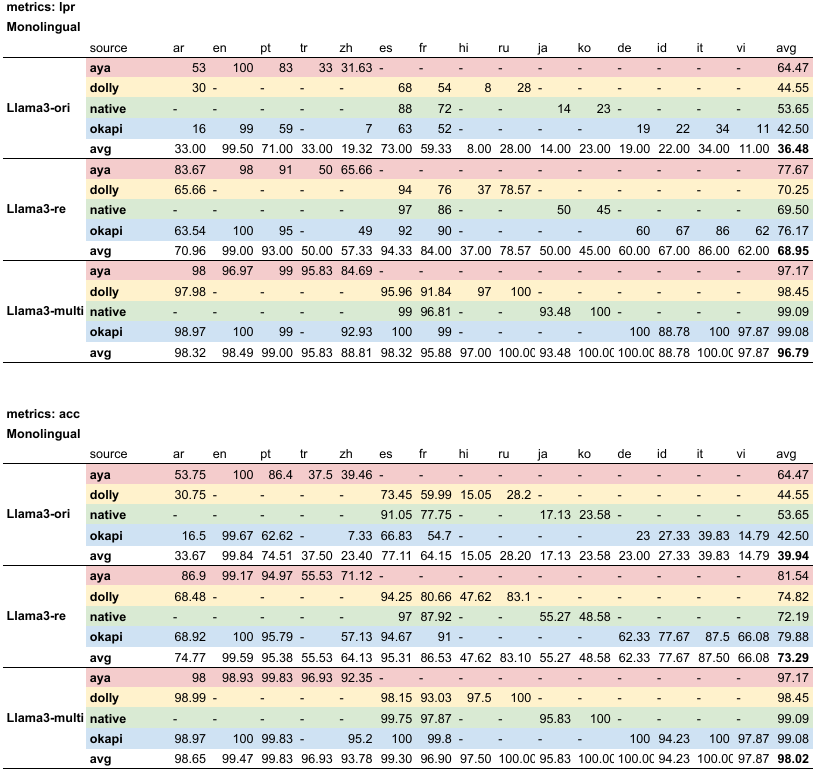}
    \caption{Full results of CP replacement experiments}
    \label{full_cp_results}
\end{table*}

\begin{table*}
\centering
\scalebox{.9}{
\begin{tabular}{lccccccccc} 
\toprule
    & \multicolumn{1}{c}{num\_ori} & \multicolumn{1}{c}{prob\_ori} & \multicolumn{1}{c}{num\_edit} & \multicolumn{1}{c}{prob\_edit} & \multicolumn{1}{c}{num\_diff} & \multicolumn{1}{c}{prob\_diff} & \multicolumn{1}{c}{fluency\_ori} & \multicolumn{1}{c}{fluency\_cna} & \multicolumn{1}{c}{diff}  \\ 
\hline
ar  & 2.83                          & 25.8                           & 5.37                           & 30.3                            & 2.55                           & 4.5                             & 30.1                              & 24.7                              & -5.4                       \\ 

pt  & 2.86                          & 49.5                           & 3.41                           & 56.0                            & 0.56                           & 6.5                             & 25.7                              & 23.3                              & -2.3                       \\ 

tr  & 2.05                          & 29.5                           & 2.42                           & 23.5                            & 0.37                           & -6.0                            & 21.2                              & 18.8                              & -2.5                       \\ 

zh  & 1.33                          & 8.6                            & 5.10                           & 37.3                            & 3.78                           & 28.7                            & 33.1                              & 26.0                              & -7.0                       \\ 

es  & 1.67                          & 26.5                           & 3.28                           & 50.3                            & 1.61                           & 23.8                            & 25.4                              & 23.2                              & -2.2                       \\ 

fr  & 2.48                          & 43.0                           & 2.91                           & 49.2                            & 0.43                           & 6.2                             & 21.2                              & 21.1                              & -0.1                       \\ 

hi  & 1.25                          & 12.0                           & 1.64                           & 13.7                            & 0.39                           & 1.8                             & 28.5                              & 22.9                              & -5.6                       \\ 

ru  & 1.09                          & 18.0                           & 3.21                           & 31.0                            & 2.12                           & 13.0                            & 23.7                              & 19.5                              & -4.2                       \\ 

de  & 2.73                          & 23.7                           & 4.45                           & 37.1                            & 1.72                           & 13.4                            & 23.8                              & 18.5                              & -5.3                       \\ 

it  & 1.33                          & 8.4                            & 2.50                           & 39.3                            & 1.17                           & 31.0                            & 25.7                              & 20.2                              & -5.5                       \\ 
\hline
avg & 1.96                          & 24.5                           & 3.43                           & 36.8                            & 1.47                           & 12.3                            & 25.8                              & 21.8                              & -4.0                       \\
\bottomrule
\end{tabular}}
\caption{Full results of robustness experiments. Perplexity is calculated to measure fluency.}
\label{token_prob}
\end{table*}

\begin{table}
\centering
\begin{tabular}{lcc} 
\toprule
\textbf{xnli }                                                         & \multicolumn{1}{c}{}         & \multicolumn{1}{c}{}           \\ 

language                                                      & \multicolumn{1}{c}{acc\_ori} & \multicolumn{1}{c}{acc\_edit}  \\ 
\hline
ar                                                            & 0.42                          & 0.37                            \\ 

de                                                            & 0.54                          & 0.54                            \\ 

es                                                            & 0.46                          & 0.5                             \\ 
fr                                                            & 0.49                          & 0.5                             \\ 
hi                                                            & 0.47                          & 0.48                            \\ 
ru                                                            & 0.37                          & 0.3                             \\ 

tr                                                            & 0.46                          & 0.52                            \\ 

vi                                                            & 0.46                          & 0.37                            \\ 

zh                                                            & 0.51                          & 0.46                            \\ 
\midrule
avg                                                           & 0.464                         & 0.449                           \\ 
\hline

\hline
& & \\
\toprule
\multicolumn{1}{c}{\textbf{sentiment analysis}} & \multicolumn{1}{c}{}         & \multicolumn{1}{c}{}           \\ 

language                                                      & \multicolumn{1}{c}{acc\_ori} & \multicolumn{1}{c}{acc\_edit}  \\ 
\hline
de                                                            & 0.98                          & 0.98                            \\ 

es                                                            & 0.98                          & 0.98                            \\ 

fr                                                            & 0.98                          & 0.97                            \\ 

ja                                                            & 0.99                          & 0.99                            \\ 

zh                                                            & 0.99                          & 0.99                            \\ 
\midrule
avg                                                           & 0.984                         & 0.982                           \\
\bottomrule
\end{tabular}
\caption{Full results of generalization experiments.}
\label{full_generalization}
\end{table}

\end{document}